\documentclass{article}

\usepackage{times}
\usepackage{graphicx} 
\usepackage{hyperref}
\usepackage{url}
\usepackage{amssymb}
\usepackage{graphicx}
\usepackage{caption}
\usepackage{subcaption}
\usepackage{sidecap}
\usepackage{slashbox}
\usepackage{natbib}

\usepackage{algorithm}
\usepackage{algorithmic}

\usepackage{hyperref}

\usepackage[accepted]{icml2016} 

\icmltitlerunning{Unsupervised Deep Embedding for Clustering Analysis}

\begin{document} 

\twocolumn[
\icmltitle{Unsupervised Deep Embedding for Clustering Analysis}

\icmlauthor{Junyuan Xie}{jxie@cs.washington.edu}
\icmladdress{University of Washington}
\icmlauthor{Ross Girshick}{rbg@fb.com}
\icmladdress{Facebook AI Research (FAIR)}
\icmlauthor{Ali Farhadi}{ali@cs.washington.edu}
\icmladdress{University of Washington}

\icmlkeywords{deep learning, machine learning}

\vskip 0.3in
]

\begin{abstract}
Clustering is central to many data-driven application domains and has been studied extensively in terms of distance functions and grouping algorithms.
Relatively little work has focused on learning representations for clustering.
In this paper, we propose Deep Embedded Clustering (DEC), a method that simultaneously learns feature representations and cluster assignments using deep neural networks.
DEC learns a mapping from the data space to a lower-dimensional feature space in which it iteratively optimizes a clustering objective.
Our experimental evaluations on image and text corpora show significant improvement over state-of-the-art methods.
\end{abstract}

\section{Introduction}
Clustering, an essential data analysis and visualization tool, has been studied
extensively in unsupervised machine learning from different perspectives: What
defines a cluster? What is the right distance metric? How to efficiently group
instances into clusters? How to validate clusters? And so on. Numerous different
distance functions and embedding methods have been explored in the literature.
Relatively little work has focused on the unsupervised learning of the feature
space in which to perform clustering.

A notion of \emph{distance} or \emph{dissimilarity} is central to data
clustering algorithms. Distance, in turn, relies on representing the data in
a feature space. The $k$-means clustering algorithm~\citep{macqueen1967some},
for example, uses the Euclidean distance between points in a given feature
space, which for images might be raw pixels or gradient-orientation histograms. The choice of feature space is customarily left as an application-specific detail for the end-user to determine. Yet it is clear that the choice of feature space is crucial; for all but the simplest image datasets, clustering with Euclidean distance on raw pixels is completely ineffective.
In this paper, we revisit cluster analysis and ask: \emph{Can we use a data driven approach to solve for the feature space and cluster memberships jointly?}

We take inspiration from recent work on deep learning for computer vision~\citep{krizhevsky2012imagenet,girshick2014rich,zeiler2014visualizing,long2014fully}, where clear gains on benchmark tasks have resulted from learning better features.
These improvements, however, were obtained with \emph{supervised} learning, whereas our goal is \emph{unsupervised} data clustering.
To this end, we define a parameterized non-linear mapping from the data space $X$ to a lower-dimensional feature space $Z$, where we optimize a clustering objective.
Unlike previous work, which operates on the data space or a shallow linear embedded space, we use stochastic gradient descent (SGD) via backpropagation on a clustering objective to learn the mapping, which is parameterized by a deep neural network.
We refer to this clustering algorithm as \emph{Deep Embedded Clustering}, or DEC.

Optimizing DEC is challenging.
We want to simultaneously solve for cluster assignment and the underlying feature representation.
However, unlike in supervised learning, we cannot train our deep network with labeled data.
Instead we propose to iteratively refine clusters with an auxiliary target distribution derived from the current soft cluster assignment.
This process gradually improves the clustering as well as the feature representation.

Our experiments show significant improvements over state-of-the-art clustering methods in terms of both accuracy and running time on image and textual datasets.
We evaluate DEC on MNIST~\citep{lecun1998gradient}, STL~\citep{coates2011analysis}, and REUTERS~\citep{lewis2004rcv1}, comparing it with standard and state-of-the-art clustering methods~\citep{nie2011spectral,yang2010image}.
In addition, our experiments show that DEC is significantly less sensitive to the choice of hyperparameters compared to state-of-the-art methods.
This robustness is an important property of our clustering algorithm since, when applied to real data, supervision is not available for hyperparameter cross-validation. 

Our contributions are:
(a) joint optimization of deep embedding and clustering;
(b) a novel iterative refinement via soft assignment;
and (c) state-of-the-art clustering results in terms of clustering accuracy and
speed.
Our Caffe-based~\citep{jia2014caffe} implementation of DEC is available at \url{https://github.com/piiswrong/dec}.

\section{Related work}
Clustering has been extensively studied in machine learning in terms of feature selection~\citep{boutsidis2009unsupervised,liu2005toward,alelyani2013feature}, distance functions~\citep{xing2002distance,xiang2008learning}, grouping methods~\citep{macqueen1967some,von2007tutorial,li2004entropy}, and cluster validation~\citep{halkidi2001clustering}.
Space does not allow for a comprehensive literature study and we refer readers to~\citep{aggarwal2013data} for a survey.

One branch of popular methods for clustering is $k$-means~\citep{macqueen1967some} and Gaussian Mixture Models (GMM)~\citep{bishop2006pattern}.
These methods are fast and applicable to a wide range of problems.
However, their distance metrics are limited to the original data space and they tend to be ineffective when input dimensionality is high~\citep{steinbach2004challenges}.

Several variants of $k$-means have been proposed to address issues with higher-dimensional input spaces.
\citet{de2006discriminative,ye2008discriminative} perform joint dimensionality reduction and clustering by first clustering the data with $k$-means and then projecting the data into a lower dimensions where the inter-cluster variance is maximized.
This process is repeated in EM-style iterations until convergence.
However, this framework is limited to linear embedding; our method employs deep neural networks to perform non-linear embedding that is necessary for more complex data.

Spectral clustering and its variants have gained popularity recently~\citep{von2007tutorial}.
They allow more flexible distance metrics and generally perform better than $k$-means.
Combining spectral clustering and embedding has been explored in \citet{yang2010image,nie2011spectral}.
\citet{tian2014learning} proposes an algorithm based on spectral clustering, but replaces eigenvalue decomposition with deep autoencoder, which improves performance but further increases memory consumption.

Most spectral clustering algorithms need to compute the full graph Laplacian matrix and therefore have quadratic or super quadratic complexities in the number of data points.
This means they need specialized machines with large memory for any dataset larger than a few tens of thousands of points.
In order to scale spectral clustering to large datasets, approximate algorithms were invented to trade off performance for speed~\citep{yan2009fast}.
Our method, however, is linear in the number of data points and scales gracefully to large datasets.

Minimizing the Kullback-Leibler (KL) divergence between a data distribution and
an embedded distribution has been used for data visualization and dimensionality
reduction~\citep{van2008visualizing}. T-SNE, for instance, is a non-parametric
algorithm in this school and a parametric variant of
t-SNE~\citep{maaten2009learning} uses deep neural network to parametrize the
embedding. The complexity of t-SNE is $O(n^2)$, where $n$ is the number of data points, but it can be approximated in $O(n\log n)$~\citep{van2014accelerating}. 

We take inspiration from parametric t-SNE. Instead of minimizing KL divergence to produce an embedding that is faithful to distances in the original data space, we define a centroid-based probability distribution and minimize its KL divergence to an auxiliary target distribution to simultaneously improve clustering assignment and feature representation. A centroid-based method also has the benefit of reducing complexity to $O(nk)$, where $k$ is the number of centroids.

\section{Deep embedded clustering}
Consider the problem of clustering a set of $n$ points $\{x_i \in X\}_{i=1}^n$ into $k$ clusters, each represented by a centroid $\mu_j, j = 1,\ldots,k$.
Instead of clustering directly in the \emph{data space} $X$, we propose to first transform the data with a non-linear mapping $f_\theta: X \rightarrow Z$, where $\theta$ are learnable parameters and $Z$ is the latent \emph{feature space}.
The dimensionality of $Z$ is typically much smaller than $X$ in order to avoid the ``curse of dimensionality''~\citep{bellman61}.
To parametrize $f_\theta$, deep neural networks (DNNs) are a natural choice due to their theoretical function approximation properties~\citep{hornik1991approximation} and their demonstrated feature learning capabilities~\citep{bengio2013representation}.

The proposed algorithm (DEC) clusters data by \emph{simultaneously} learning a set of $k$ cluster centers $\{\mu_j \in Z\}_{j=1}^{k}$ in the feature space $Z$ and the parameters $\theta$ of the DNN that maps data points into $Z$. DEC has two phases: (1) parameter initialization with a deep autoencoder~\citep{vincent2010stacked} and (2) parameter optimization (i.e., clustering), where we iterate between computing an auxiliary target distribution and minimizing the Kullback--Leibler (KL) divergence to it. We start by describing phase (2) parameter optimization/clustering, given an initial estimate of $\theta$ and $\{\mu_j\}_{j=1}^{k}$.

\subsection{Clustering with KL divergence}
Given an initial estimate of the non-linear mapping $f_\theta$ and the initial cluster centroids $\{\mu_j\}_{j=1}^{k}$, we propose to improve the clustering using an unsupervised algorithm that alternates between two steps.
In the first step, we compute a soft assignment between the embedded points and the cluster centroids. In the second step, we update the deep mapping $f_\theta$ and refine the cluster centroids by learning from current high confidence assignments using an auxiliary target distribution.
This process is repeated until a convergence criterion is met.

\subsubsection{Soft Assignment}
Following \citet{van2008visualizing} we use the Student's $t$-distribution as a kernel to measure the similarity between embedded point $z_i$ and centroid $\mu_j$:
\begin{equation}
q_{ij} = \frac{(1+\Vert z_i - \mu_j\Vert^2/\alpha)^{-\frac{\alpha+1}{2}}}{\sum_{j'} (1+\Vert z_i - \mu_{j'}\Vert^2/\alpha)^{-\frac{\alpha+1}{2}}},
\end{equation}
where $z_i = f_\theta(x_i) \in Z$ corresponds to $x_i \in X$ after embedding,  $\alpha$ are the degrees of freedom of the Student's $t$-distribution and $q_{ij}$ can be interpreted as the probability of assigning sample $i$ to cluster $j$ (i.e., a soft assignment). Since we cannot cross-validate $\alpha$ on a validation set in the unsupervised setting, and learning it is superfluous~\citep{maaten2009learning}, we let $\alpha = 1$ for all experiments.

\subsubsection{KL divergence minimization}
We propose to iteratively refine the clusters by learning from their high confidence assignments with the help of an auxiliary target distribution. 
Specifically, our model is trained by matching the soft assignment to the target distribution.
To this end, we define our objective as a KL divergence loss between the soft assignments $q_i$ and the auxiliary distribution $p_i$ as follows:
\begin{equation}
L = \mathrm{KL}(P\Vert Q) = \sum_i \sum_j p_{ij}\log \frac{p_{ij}}{q_{ij}}.
\end{equation}
The choice of target distributions $P$ is crucial for DEC's performance. A naive
approach would be setting each $p_{i}$ to a delta distribution (to the nearest
centroid) for data points above a confidence threshold and ignore the rest.
However, because $q_i$ are soft assignments, it is more natural and flexible to
use softer probabilistic targets. Specifically, we would like our target distribution to
have the following properties: (1) strengthen predictions (i.e., improve cluster
purity), (2) put more emphasis on data points assigned with high confidence, and
(3) normalize loss contribution of each centroid to prevent large clusters from distorting the hidden feature space.

In our experiments, we compute $p_i$ by first raising $q_i$ to the second power and then normalizing by frequency per cluster:
\begin{equation}
p_{ij} = \frac{q_{ij}^2/f_j}{\sum_{j'} q_{ij'}^2/f_{j'}},
\end{equation}
where $f_j = \sum_i q_{ij}$ are soft cluster frequencies. Please refer to section~\ref{sec:objective} for discussions on empirical properties of $L$ and $P$.

Our training strategy can be seen as a form of self-training~\cite{nigam2000analyzing}. As in self-training, we take an initial classifier and an unlabeled dataset, then label the dataset with the classifier in order to train on its own high confidence predictions. Indeed, in experiments we observe that DEC improves upon the initial estimate in each iteration by learning from high confidence predictions, which in turn helps to improve low confidence ones.

\subsubsection{Optimization}
We jointly optimize the cluster centers $\{\mu_j\}$ and DNN parameters $\theta$ using Stochastic Gradient Descent (SGD) with momentum. The gradients of $L$ with respect to feature-space embedding of each data point $z_i$ and each cluster centroid $\mu_j$ are computed as:
\begin{eqnarray}
\frac{\partial L}{\partial z_i} &=& \frac{\alpha+1}{\alpha}\sum_j(1+\frac{\Vert z_i - \mu_j\Vert^2}{\alpha})^{-1}\\
&&\;\;\;\;\times(p_{ij}-q_{ij})(z_i-\mu_j),\nonumber\\
\frac{\partial L}{\partial \mu_j} &=& -\frac{\alpha+1}{\alpha}\sum_i(1+\frac{\Vert z_i - \mu_j\Vert^2}{\alpha})^{-1}\\
&&\;\;\;\;\times(p_{ij}-q_{ij})(z_i-\mu_j).\nonumber
\end{eqnarray}
The gradients $\partial L / \partial z_i$ are then passed down to the DNN and used in standard backpropagation to compute the DNN's parameter gradient $\partial L / \partial \theta$.
For the purpose of discovering cluster assignments, we stop our procedure when less than $\mathit{tol}\%$ of points change cluster assignment between two consecutive iterations.

\begin{table*}[!t]
\centering
\caption{Dataset statistics.}
\begin{tabular}{l|c|c|c|c} 
Dataset  & \# Points & \# classes & Dimension & \% of largest class \\ \hline \hline
MNIST \cite{lecun1998gradient}   & 70000     & 10         & 784       & 11\%                \\ \hline
STL-10 \cite{coates2011analysis} & 13000     & 10         & 1428      & 10\%                \\ \hline
REUTERS-10K & 10000     & 4          & 2000      & 43\%                \\ \hline
REUTERS \cite{lewis2004rcv1} & 685071     & 4          & 2000      & 43\%                \\
\end{tabular}
\label{table:dataset}
\end{table*}

\subsection{Parameter initialization}
\begin{figure}[t!]
\centering
\includegraphics[width=0.8\linewidth]{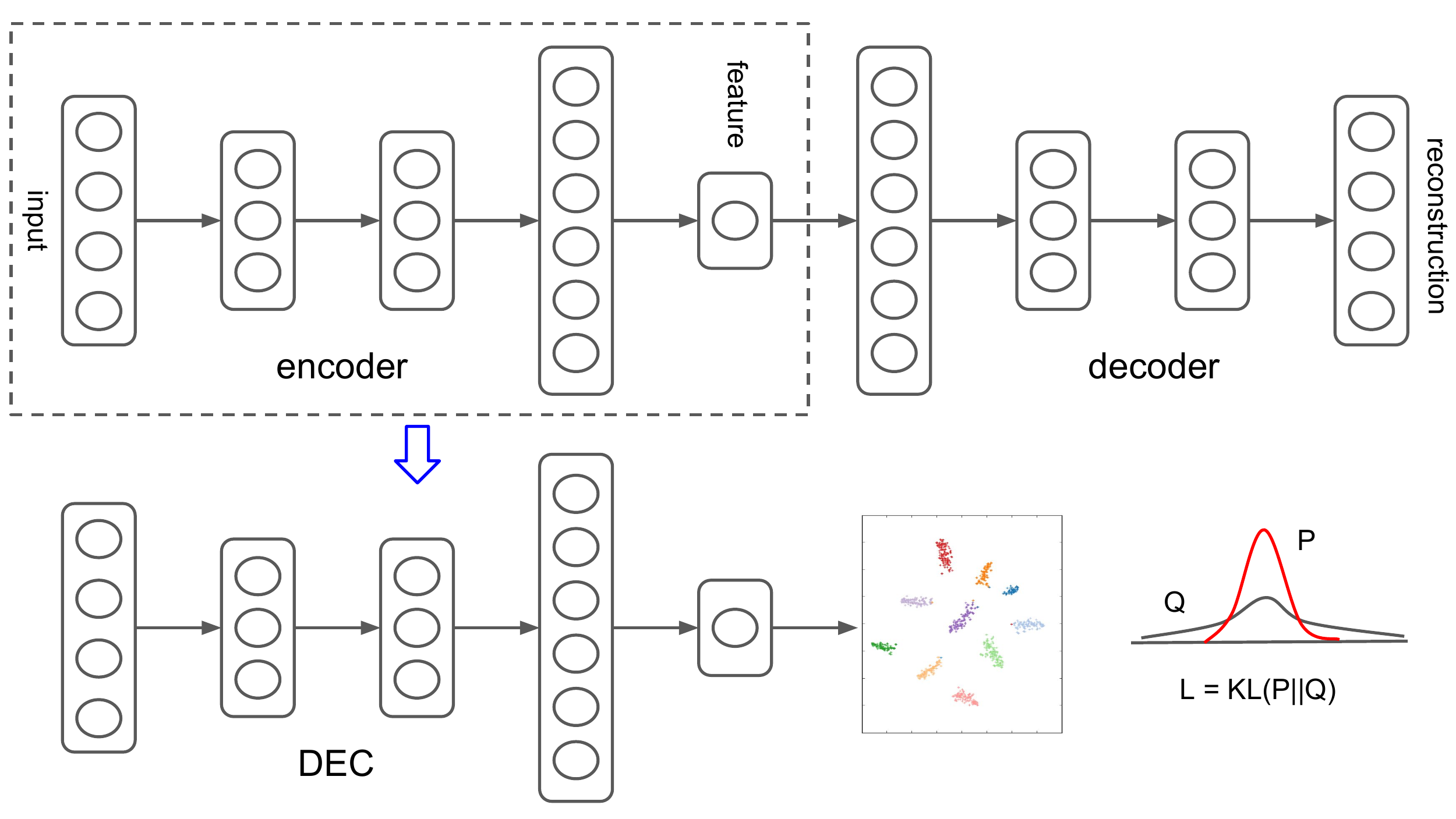}
\caption{Network structure}
\label{fig:network}
\end{figure}

Thus far we have discussed how DEC proceeds given initial estimates of the DNN parameters $\theta$ and the cluster centroids $\{\mu_j\}$.
Now we discuss how the parameters and centroids are initialized.

We initialize DEC with a stacked autoencoder (SAE) because recent research has shown that they consistently produce semantically meaningful and well-separated representations on real-world datasets~\citep{vincent2010stacked,hinton2006reducing,le2013building}. Thus the unsupervised representation learned by SAE naturally facilitates the learning of clustering representations with DEC.

We initialize the SAE network layer by layer with each layer being a denoising autoencoder
trained to reconstruct the previous layer's output after random
corruption~\citep{vincent2010stacked}. A denoising autoencoder is a two layer neural network defined as:
\begin{eqnarray}
  \tilde x \sim \mathit{Dropout}(x)\\
h = g_1(W_1\tilde x+b_1)\\
\tilde h \sim \mathit{Dropout}(h)\\
y = g_2(W_2\tilde h + b_2)
\end{eqnarray}
where $\mathit{Dropout}(\cdot)$~\citep{srivastava2014dropout} is a stochastic mapping that randomly sets a portion of its input dimensions to 0, $g_1$ and $g_2$ are activation functions for encoding and decoding layer respectively, and $\theta=\{W_1, b_1, W_2, b_2\}$ are model parameters. Training is performed by minimizing the least-squares loss $\Vert x - y \Vert_2^2$. After training of one layer, we use its output $h$ as the input to train the next layer.
We use rectified linear units (ReLUs)~\citep{nair2010rectified} in all encoder/decoder pairs, except for $g_2$ of the \emph{first} pair (it needs to reconstruct input data that may have positive and negative values, such as zero-mean images) and $g_1$ of the \emph{last} pair (so the final data embedding retains full information~\citep{vincent2010stacked}).

After greedy layer-wise training, we concatenate all encoder layers followed by all decoder layers, in reverse layer-wise training order, to form a deep autoencoder and then finetune it to minimize reconstruction loss. The final result is a multilayer deep autoencoder with a bottleneck coding layer in the middle. We then discard the decoder layers and use the encoder layers as our initial mapping between the data space and the feature space, as shown in Fig.~\ref{fig:network}.

To initialize the cluster centers, we pass the data through the initialized DNN to get embedded data points and then perform standard $k$-means clustering in the feature space $Z$ to obtain $k$ initial centroids $\{\mu_j\}_{j=1}^k$.

\section{Experiments}

\begin{figure*}[!ht]
\includegraphics[width=\textwidth]{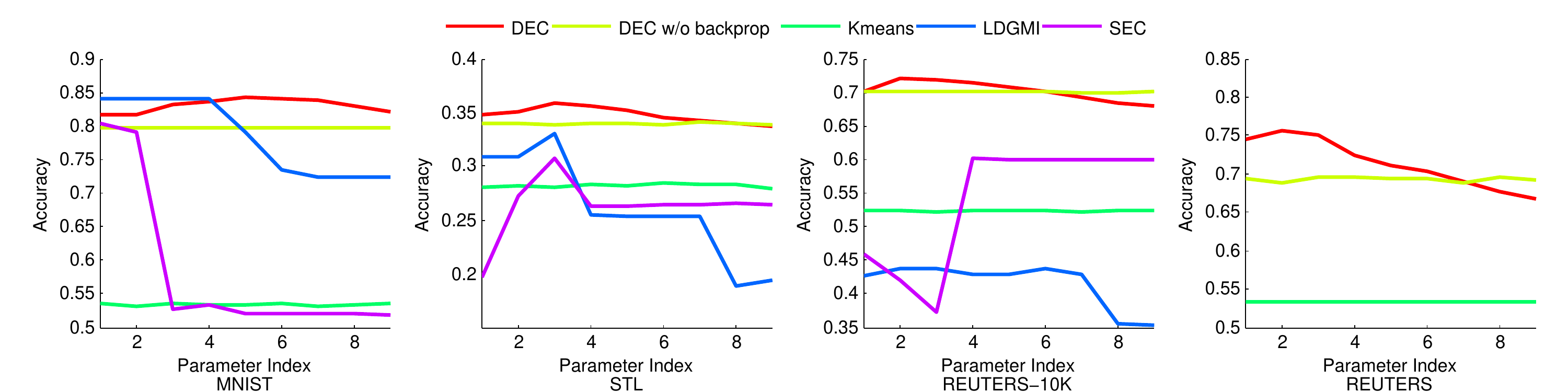}
\caption{Clustering accuracy for different hyperparameter choices for each algorithm.
DEC outperforms other methods and is more robust to hyperparameter changes compared to either LDGMI or SEC.
Robustness is important because cross-validation is not possible in real-world applications of cluster analysis. This figure is best viewed in color.}
\label{fig:acc}
\end{figure*}

\begin{table*}[!ht]
\centering
\caption{Comparison of clustering accuracy (Eq. \ref{eqn:acc}) on four datasets.}
\begin{tabular}{l|c|c|c|c}
Method	& MNIST	& STL-HOG	& REUTERS-10k	& REUTERS \\ \hline \hline
$k$-means	& 53.49\%	& 28.39\%	& 52.42\%	& 53.29\% \\ \hline
LDMGI	& 84.09\%	& 33.08\%	& 43.84\%	& N/A \\ \hline
SEC		& 80.37\%	& 30.75\%	& 60.08\%	& N/A \\ \hline
DEC w/o backprop & 79.82\% & 34.06\% & 70.05\% & 69.62\% \\ \hline
DEC (ours)	& \textbf{84.30\%}	& \textbf{35.90\%}	& \textbf{72.17\%}	& \textbf{75.63\%}
\end{tabular}
\label{table:acc}
\end{table*}

\subsection{Datasets}
We evaluate the proposed method (DEC) on one text dataset and two image datasets and compare it against other algorithms including $k$-means, LDGMI~\citep{yang2010image}, and SEC~\citep{nie2011spectral}.
LDGMI and SEC are spectral clustering based algorithms that use a Laplacian matrix and various transformations to improve clustering performance.
Empirical evidence reported in \citet{yang2010image,nie2011spectral} shows that LDMGI and SEC outperform traditional spectral clustering methods on a wide range of datasets.
We show qualitative and quantitative results that demonstrate the benefit of DEC compared to LDGMI and SEC.

In order to study the performance and generality of different algorithms, we perform experiment on two image datasets and one text data set:
\begin{itemize}
\item \textbf{MNIST}: The MNIST dataset consists of 70000 handwritten digits of 28-by-28 pixel size. The digits are centered and size-normalized~\citep{lecun1998gradient}.
\item \textbf{STL-10}: A dataset of 96-by-96 color images. There are 10 classes with 1300 examples each. It also contains 100000 unlabeled images of the same resolution~\citep{coates2011analysis}. We also used the unlabeled set when training our autoencoders. Similar to \citet{doersch2012makes}, we concatenated HOG feature and a 8-by-8 color map to use as input to all algorithms.
\item \textbf{REUTERS}: Reuters contains about 810000 English news stories labeled with a category tree~\citep{lewis2004rcv1}. We used the four root categories: corporate/industrial, government/social, markets, and economics as labels and further pruned all documents that are labeled by multiple root categories to get 685071 articles. We then computed tf-idf features on the 2000 most frequently occurring word stems. Since some algorithms do not scale to the full Reuters dataset, we also sampled a random subset of 10000 examples, which we call REUTERS-10k, for comparison purposes.
\end{itemize}
A summary of dataset statistics is shown in Table \ref{table:dataset}.
For all algorithms, we normalize all datasets so that $\frac{1}{d}\Vert x_i \Vert_2^2$ is approximately 1, where $d$ is the dimensionality of the data space point $x_i \in X$.

\subsection{Evaluation Metric}
We use the standard unsupervised evaluation metric and protocols for evaluations and comparisons to other algorithms \cite{yang2010image}.
For all algorithms we set the number of clusters to the number of ground-truth categories and evaluate performance with \emph{unsupervised clustering accuracy ($\mathit{ACC}$)}:
\begin{equation}\label{eqn:acc}
  \mathit{ACC} = \max_m \frac{\sum_{i=1}^n \mathbf{1}\{l_i = m(c_i)\}}{n},
\end{equation}
where $l_i$ is the ground-truth label, $c_i$ is the cluster assignment produced by the algorithm, and $m$ ranges over all possible one-to-one mappings between clusters and labels.

Intuitively this metric takes a cluster assignment from an \emph{unsupervised}
algorithm and a ground truth assignment and then finds the best matching between them.
The best mapping can be efficiently computed by the Hungarian algorithm~\citep{kuhn1955hungarian}.

\subsection{Implementation}

\begin{figure*}[!ht]
\centering
\begin{subfigure}[t]{0.40\textwidth}
\centering
\includegraphics[width=\linewidth]{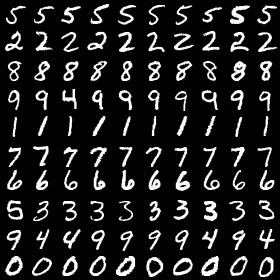}
\caption{MNIST}
\end{subfigure}
~\hspace{1cm}
\begin{subfigure}[t]{0.40\textwidth}
\centering
\includegraphics[width=\linewidth]{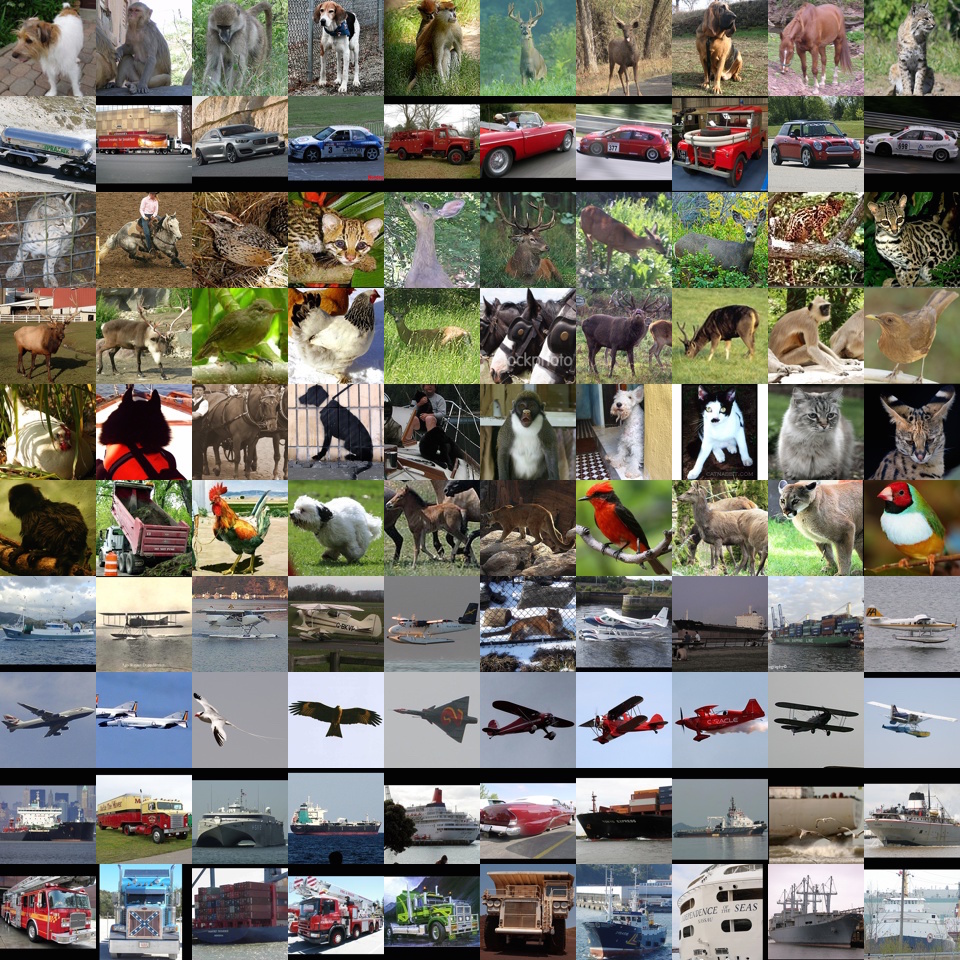}
\caption{STL-10}
\end{subfigure}
\caption{Each row contains the top 10 scoring elements from one cluster.}
\label{fig:top10}
\end{figure*}

Determining hyperparameters by cross-validation on a validation set is not an option in unsupervised clustering.
Thus we use commonly used parameters for DNNs and avoid dataset specific tuning as much as possible.
Specifically, inspired by \citet{maaten2009learning}, we set network dimensions to $d$--500--500--2000--10 for all datasets, where $d$ is the data-space dimension, which varies between datasets.
All layers are densely (fully) connected.

During greedy layer-wise pretraining we initialize the weights to random numbers drawn from a zero-mean Gaussian distribution with a standard deviation of 0.01.
Each layer is pretrained for 50000 iterations with a dropout rate of $20\%$.
The entire deep autoencoder is further finetuned for 100000 iterations without dropout.
For both layer-wise pretraining and end-to-end finetuning of the autoencoder the minibatch size is set to 256, starting learning rate is set to 0.1, which is divided by 10 every 20000 iterations, and weight decay is set to 0.
All of the above parameters are set to achieve a reasonably good reconstruction loss and are held constant across all datasets.
Dataset-specific settings of these parameters might improve performance on each dataset, but we refrain from this type of unrealistic parameter tuning.
To initialize centroids, we run $k$-means with 20 restarts and select the best solution.
In the KL divergence minimization phase, we train with a constant learning rate of 0.01.
The convergence threshold is set to $\mathit{tol} = 0.1\%$.
Our implementation is based on Python and Caffe~\citep{jia2014caffe} and is available at \url{https://github.com/piiswrong/dec}.

For all baseline algorithms, we perform 20 random restarts when initializing centroids and pick the result with the best objective value.
For a fair comparison with previous work~\citep{yang2010image}, we vary one hyperparameter for each algorithm over 9 possible choices and report the best accuracy in Table \ref{table:acc} and the range of accuracies in Fig. \ref{fig:acc}.
For LDGMI and SEC, we use the same parameter and range as in their corresponding papers.
For our proposed algorithm, we vary $\lambda$, the parameter that controls annealing speed, over $2^i\times 10, i = 0, 1, ..., 8$.
Since $k$-means does not have tunable hyperparameters (aside from $k$), we simply run them 9 times.
GMMs perform similarly to $k$-means so we only report $k$-means results. Traditional spectral clustering performs worse than LDGMI and SEC so we only report the latter~\citep{yang2010image,nie2011spectral}.

\subsection{Experiment results}

We evaluate the performance of our algorithm both quantitatively and qualitatively.
In Table \ref{table:acc}, we report the best performance, over 9 hyperparameter settings, of each algorithm.
Note that DEC outperforms all other methods, sometimes with a significant margin.
To demonstrate the effectiveness of end-to-end training, we also show the
results from freezing the non-linear mapping $f_\theta$ during clustering.
We find that this ablation (``DEC w/o backprop'') generally performs worse than DEC. 

In order to investigate the effect of hyperparameters, we plot the accuracy of
each method under all 9 settings (Fig. \ref{fig:acc}).
We observe that DEC is more consistent across hyperparameter ranges compared to LDGMI and SEC.
For DEC, hyperparameter $\lambda = 40$ gives near optimal performance on all dataset, whereas for other algorithms the optimal hyperparameter varies widely.
Moreover, DEC can process the entire REUTERS dataset in half an hour with GPU acceleration while the second best algorithms, LDGMI and SEC, would need months of computation time and terabytes of memory.
We, indeed, could not run these methods on the full REUTERS dataset and report
N/A in Table \ref{table:acc} (GPU adaptation of these methods is non-trivial).

In Fig. \ref{fig:top10} we show 10 top scoring images from each cluster in MNIST and STL.
Each row corresponds to a cluster and images are sorted from left to right based on their distance to the cluster center.
We observe that for MNIST, DEC's cluster assignment corresponds to natural clusters very well, with the exception of confusing 4 and 9, while for STL, DEC is mostly correct with airplanes, trucks and cars, but spends part of its attention on poses instead of categories when it comes to animal classes.

\section{Discussion}
\subsection{Assumptions and Objective}

\begin{figure}[!h]
\centering
\includegraphics[width=0.45\textwidth]{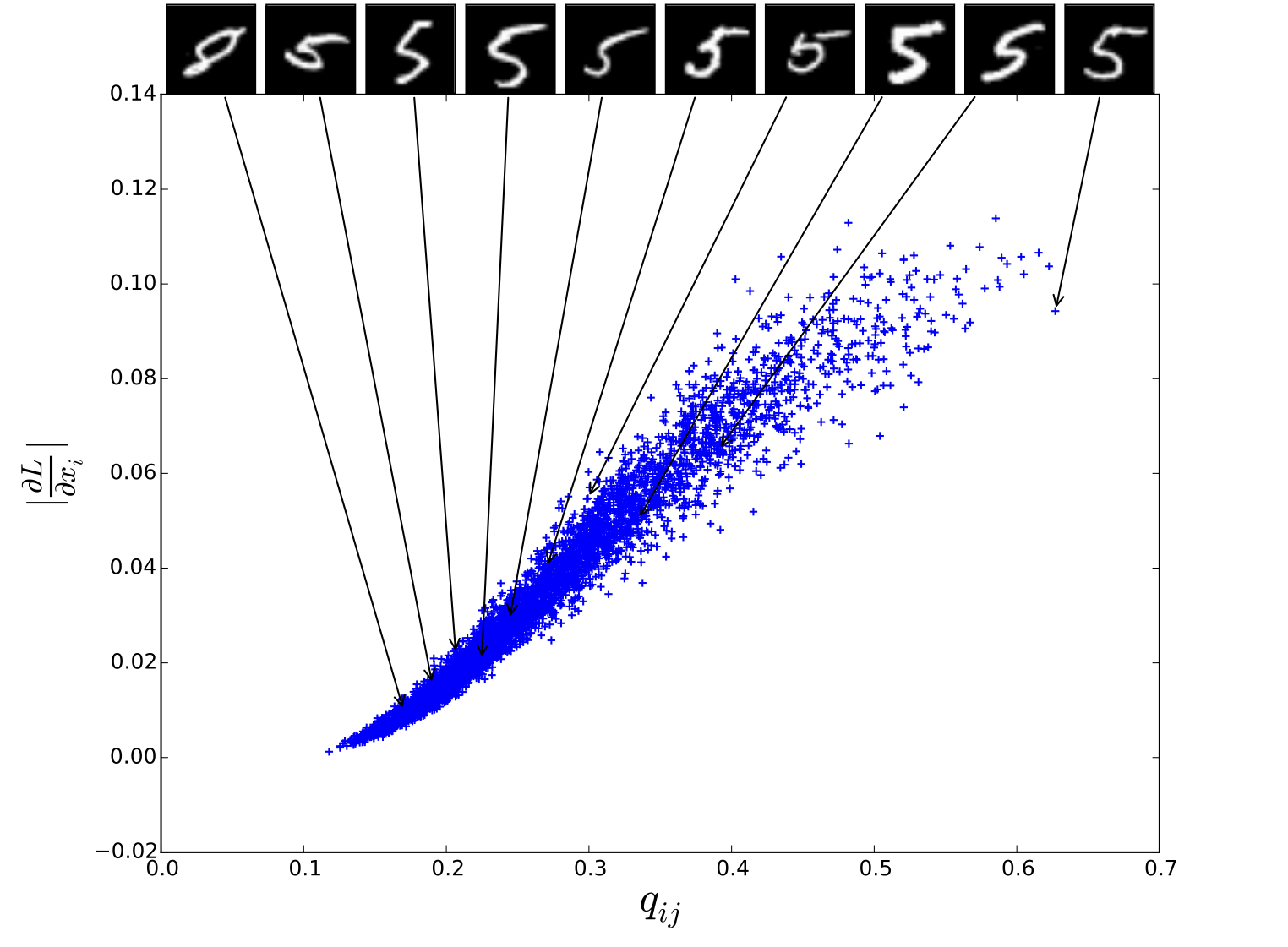}
\caption{Gradient visualization at the start of KL divergence minimization.
This plot shows the magnitude of the gradient of the loss $L$ vs. the cluster soft assignment probability $q_{ij}$.
See text for discussion.}
\label{fig:grad}
\end{figure}
\label{sec:objective}

\begin{figure*}[t]
\centering
\begin{subfigure}[b]{0.25\textwidth}
\includegraphics[width=\textwidth]{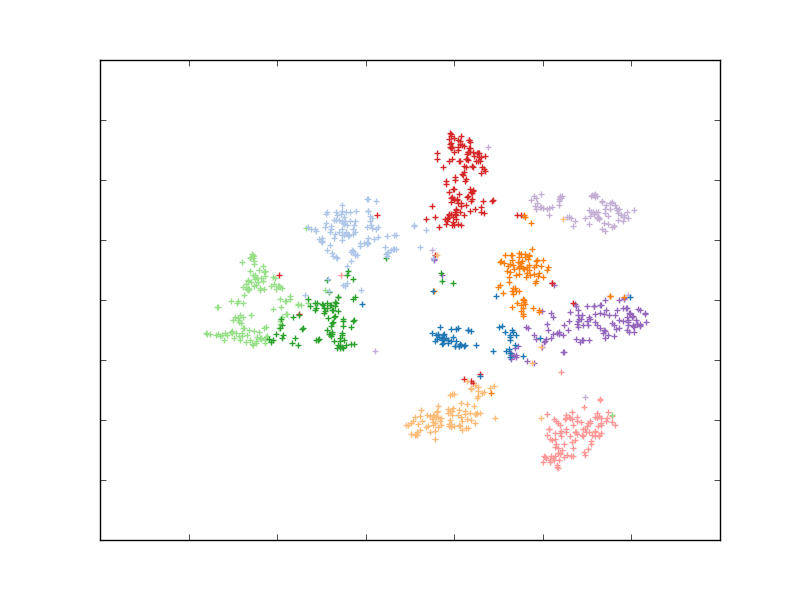}
\caption{Epoch 0}
\end{subfigure}\quad\quad
\begin{subfigure}[b]{0.25\textwidth}
\includegraphics[width=\textwidth]{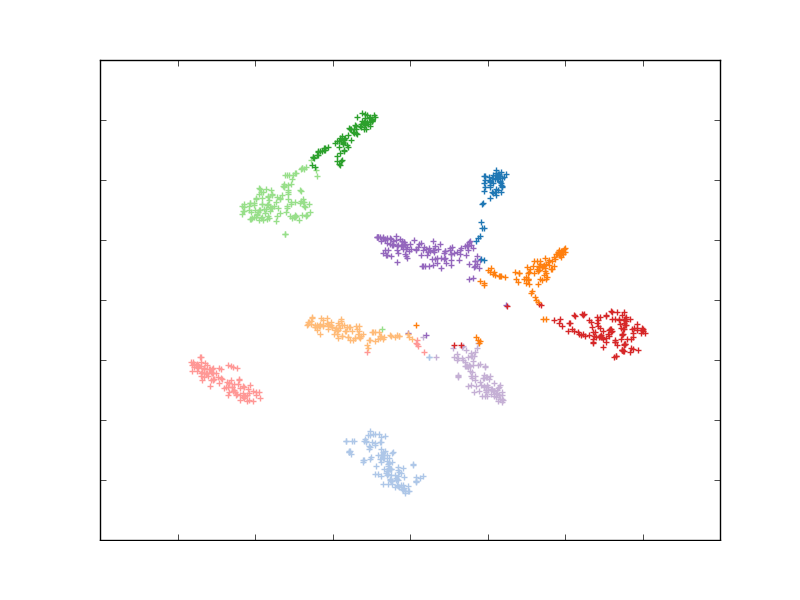}
\caption{Epoch 3}
\end{subfigure}\quad\quad
\begin{subfigure}[b]{0.25\textwidth}
\includegraphics[width=\textwidth]{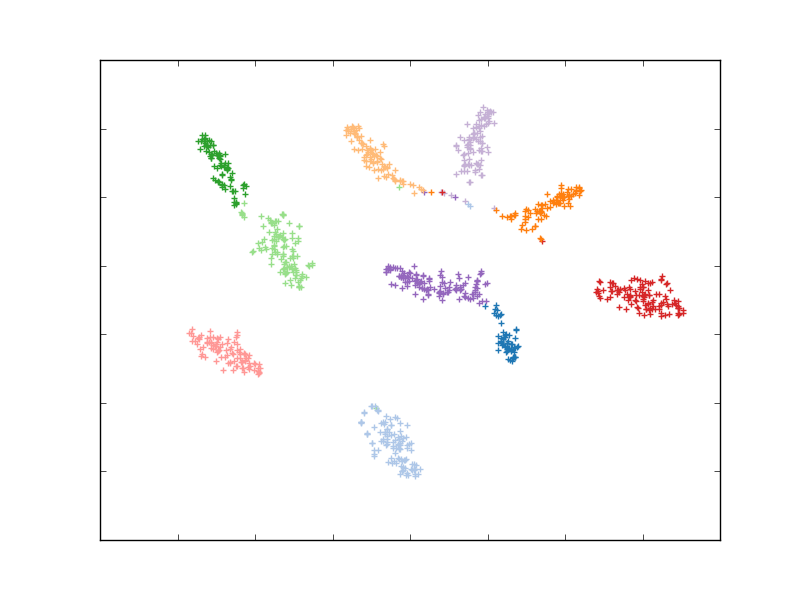}
\caption{Epoch 6}
\end{subfigure}\\
\begin{subfigure}[b]{0.25\textwidth}
\includegraphics[width=\textwidth]{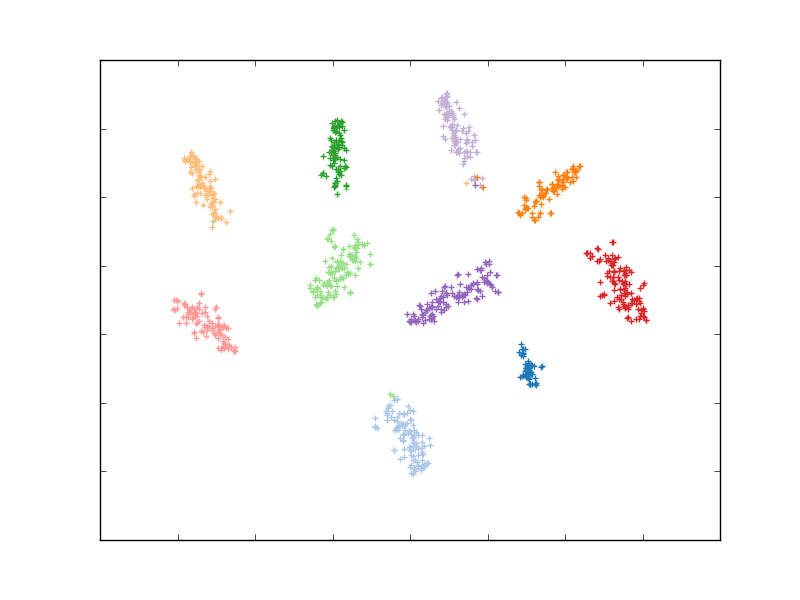}
\caption{Epoch 9}
\end{subfigure}\quad\quad
\begin{subfigure}[b]{0.25\textwidth}
\includegraphics[width=\textwidth]{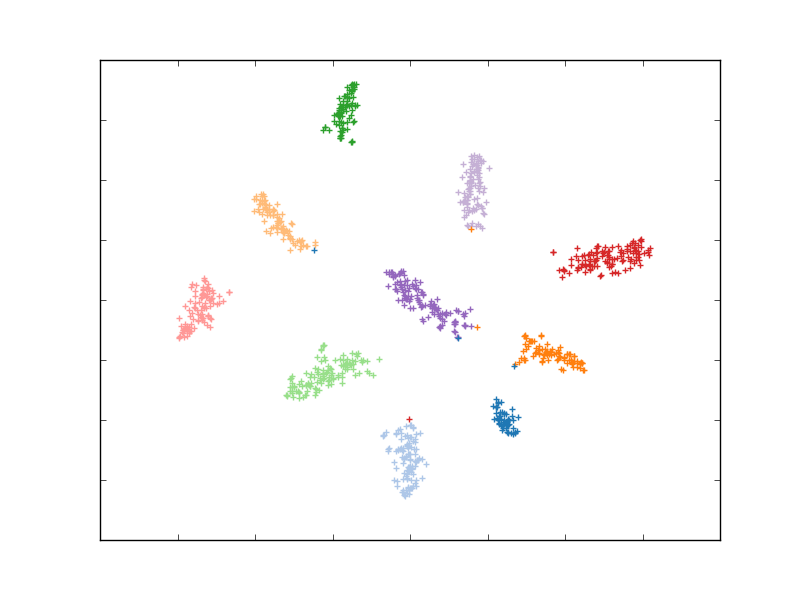}
\caption{Epoch 12}
\end{subfigure}\quad\quad
\begin{subfigure}[b]{0.25\textwidth}
\includegraphics[width=\textwidth]{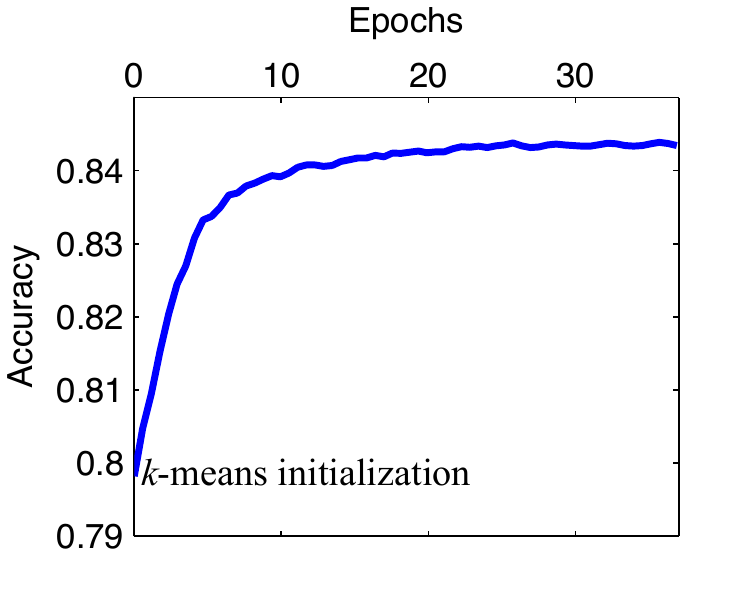}
\caption{Accuracy vs. epochs}
\end{subfigure}
\caption{We visualize the latent representation as the KL divergence minimization phase proceeds on MNIST.
Note the separation of clusters from epoch 0 to epoch 12.
We also plot the accuracy of DEC at different epochs, showing that KL divergence minimization improves clustering accuracy. This figure is best viewed in color.}
\label{fig:progress}
\end{figure*}

\begin{table*}[ht]
\centering
\caption{Comparison of clustering accuracy (Eq. \ref{eqn:acc}) on autoencoder (AE) feature.}
\begin{tabular}{l|c|c|c|c}
Method	& MNIST	& STL-HOG	& REUTERS-10k	& REUTERS \\ \hline \hline
AE+$k$-means	& 81.84\%	& 33.92\%	& 66.59\%	& 71.97\% \\ \hline
AE+LDMGI	& 83.98\%	& 32.04\%	& 42.92\%	& N/A \\ \hline
AE+SEC		& 81.56\%	& 32.29\%	& 61.86\%	& N/A \\ \hline
DEC (ours)	& \textbf{84.30\%}	& \textbf{35.90\%}	& \textbf{72.17\%}	& \textbf{75.63\%}
\end{tabular}
\label{table:ae}
\end{table*}

\begin{table*}[ht]
\centering
\caption{Clustering accuracy (Eq. \ref{eqn:acc}) on imbalanced subsample of MNIST.}
\begin{tabular}{l|c|c|c|c|c} 
\backslashbox{Method}{$r_{min}$}  & 0.1 & 0.3 & 0.5 & 0.7 & 0.9 \\ \hline \hline
$k$-means & 47.14\%     & 49.93\%         & 53.65\%       & 54.16\%  &  54.39\%            \\ \hline
AE+$k$-means & 66.82\%     & 74.91\%         & 77.93\%      & 80.04\%  &   81.31\%           \\ \hline
DEC & 70.10\%     & 80.92\%          & 82.68\%      & 84.69\%  &  85.41\%            \\ 
\end{tabular}
\label{table:imba}
\end{table*}

The underlying assumption of DEC is that the initial classifier's high confidence predictions are mostly correct.
To verify that this assumption holds for our task and that our choice of $P$ has the desired properties, we plot the magnitude of the gradient of $L$ with respect to each embedded point, $|\partial L / \partial z_i|$, against its soft assignment, $q_{ij}$, to a randomly chosen MNIST cluster $j$ (Fig. \ref{fig:grad}).

We observe points that are closer to the cluster center (large $q_{ij}$) contribute more to the gradient.
We also show the raw images of 10 data points at each 10 percentile sorted by $q_{ij}$.
Instances with higher similarity are more canonical examples of ``5''.
As confidence decreases, instances become more ambiguous and eventually turn into a mislabeled ``8'' suggesting the soundness of our assumptions.

\subsection{Contribution of Iterative Optimization}

In Fig. \ref{fig:progress} we visualize the progression of the embedded representation of a random subset of MNIST during training.
For visualization we use t-SNE~\citep{van2008visualizing} applied to the embedded points $z_i$.
It is clear that the clusters are becoming increasingly well separated.
Fig. \ref{fig:progress}  (f) shows how accuracy correspondingly improves over SGD epochs.

\subsection{Contribution of Autoencoder Initialization}
To better understand the contribution of each component, we show the performance of all algorithms with autoencoder features in Table \ref{table:ae}.
We observe that SEC and LDMGI's performance do not change significantly with autoencoder feature, while $k$-means improved but is still below DEC.
This demonstrates the power of deep embedding and the benefit of fine-tuning with the proposed KL divergence objective.

\subsection{Performance on Imbalanced Data}
In order to study the effect of imbalanced data, we sample subsets of MNIST with various retention rates.
For minimum retention rate $r_{min}$, data points of class 0 will be kept with probability $r_{min}$ and class 9 with probability 1, with the other classes linearly in between.
As a result the largest cluster will be $1/r_{min}$ times as large as the smallest one. 
From Table \ref{table:imba} we can see that DEC is fairly robust against cluster size variation.
We also observe that KL divergence minimization (DEC) consistently improves clustering accuracy after autoencoder and $k$-means initialization (shown as AE+$k$-means).

\subsection{Number of Clusters}
\begin{figure}[!h]
\centering
\includegraphics[width=0.4\textwidth]{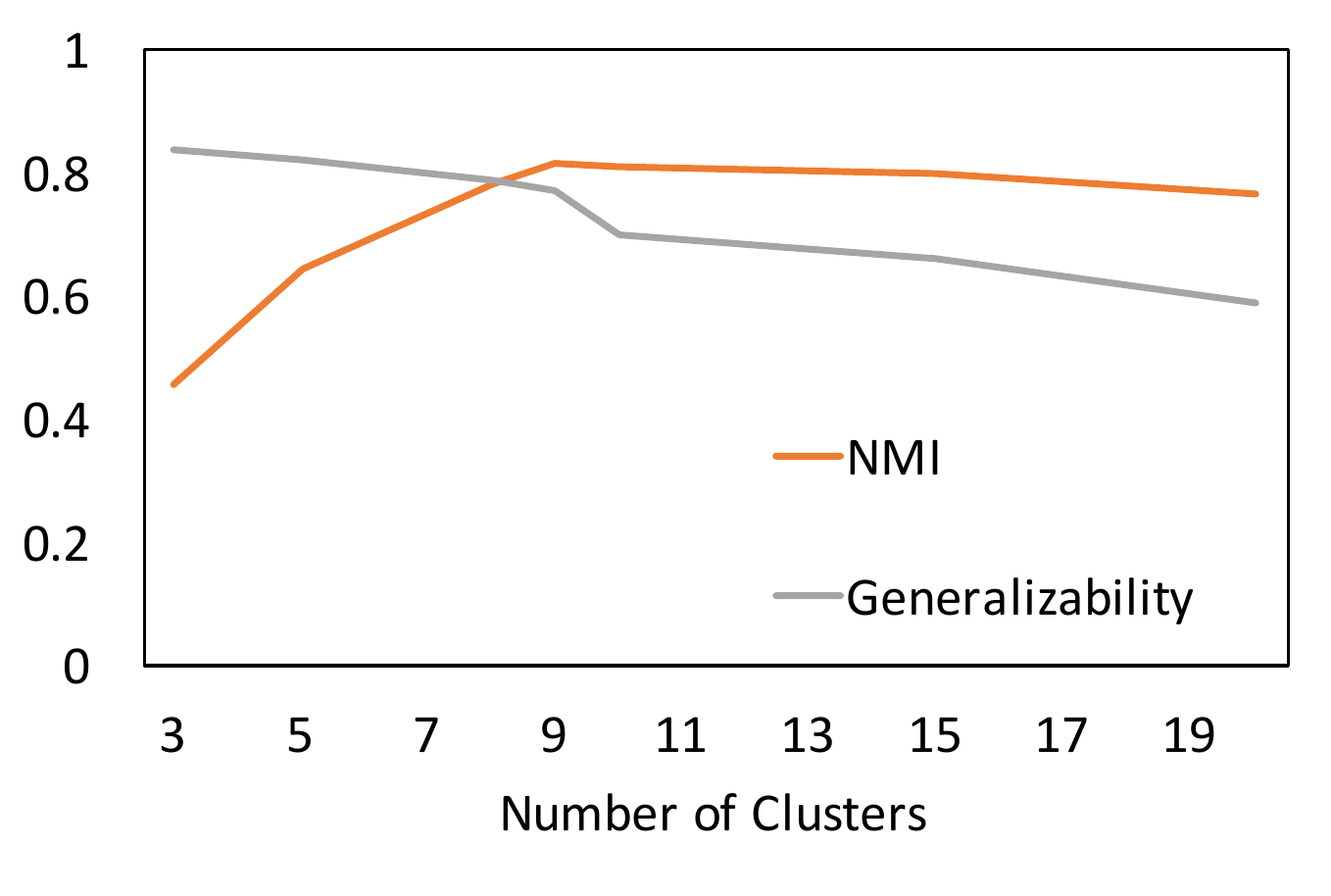}
\caption{Selection of the centroid count, $k$.
This is a plot of Normalized Mutual Information (NMI) and Generalizability vs. number of clusters.
Note that there is a sharp drop of generalizability from 9 to 10 which means that 9 is the optimal number of clusters.
Indeed, we observe that 9 gives the highest NMI.}
\label{fig:nc}
\end{figure}

So far we have assumed that the number of natural clusters is given to simplify comparison between algorithms.
However, in practice this quantity is often unknown.
Therefore a method for determining the optimal number of clusters is needed.
To this end, we define two metrics: (1) the standard metric, Normalized Mutual Information (NMI), for evaluating clustering results with different cluster number:
\begin{displaymath}
  \mathit{NMI}(l, c) = \frac{I(l, c)}{\frac{1}{2}[H(l)+H(c)]},
\end{displaymath}
where $I$ is the mutual information metric and $H$ is entropy,
and (2) generalizability ($G$) which is defined as the ratio between training and validation loss:
\begin{displaymath}
G = \frac{L_{train}}{L_{validation}}.
\end{displaymath}
$G$ is small when training loss is lower than validation loss, which indicate a high degree of overfitting.

Fig. \ref{fig:nc} shows a sharp drop in generalizability when cluster number increases from 9 to 10, which suggests that 9 is the optimal number of clusters. We indeed observe the highest NMI score at 9, which demonstrates that generalizability is a good metric for selecting cluster number. NMI is highest at 9 instead 10 because 
9 and 4 are similar in writing and DEC thinks that they should form a single cluster. This corresponds well with our qualitative results in Fig. \ref{fig:top10}.

\section{Conclusion}
This paper presents Deep Embedded Clustering, or DEC---an algorithm that clusters a set of data points in a jointly optimized feature space.
DEC works by iteratively optimizing a KL divergence based clustering objective with a self-training target distribution. Our method can be viewed as an unsupervised extension of semisupervised self-training. Our framework provide a way to learn a representation specialized for clustering without groundtruth cluster membership labels.

Empirical studies demonstrate the strength of our proposed algorithm. DEC offers improved performance as well as robustness with respect to hyperparameter settings, which is particularly important in unsupervised tasks since cross-validation is not possible. DEC also has the virtue of linear complexity in the number of data points which allows it to scale to large datasets.

\section{Acknowledgment}
This work is in part supported by ONR N00014-13-1-0720, NSF IIS- 1338054, and Allen Distinguished Investigator Award.

\bibliography{bib}
\bibliographystyle{icml2016}

\end{document}